\documentclass[11pt]{article}

\usepackage[final]{acl}

\usepackage{times}
\usepackage{latexsym}

\usepackage[T1]{fontenc}

\usepackage[utf8]{inputenc}

\usepackage{microtype}

\usepackage{inconsolata}

\usepackage{multirow}
\usepackage {comment}
\usepackage{float}
\usepackage{amssymb}
\usepackage{amsmath}
\usepackage[table]{xcolor} 
\usepackage{subcaption}
\usepackage{tcolorbox}

\title{Investigating Training and Generalization\\in Faithful Self-Explanations of Large Language Models}

\author{
  \textbf{Tomoki Doi\textsuperscript{1,2}},
  \textbf{Masaru Isonuma\textsuperscript{1,2,3,4}},
  \textbf{Hitomi Yanaka\textsuperscript{1,2,3}}\\
  \textsuperscript{1}The University of Tokyo 
  \textsuperscript{2}Riken 
  \textsuperscript{3}Tohoku University
  \textsuperscript{4}NII LLMC\\
  \texttt{\{doi-tomoki701, hyanaka\}@is.s.u-tokyo.ac.jp}\\
  \texttt{isonuma@nii.ac.jp}
}

\begin{document}
\maketitle
\begin{abstract}
Large language models have the potential to generate explanations for their own predictions in a variety of styles based on user instructions.
Recent research has examined whether these self-explanations faithfully reflect the models' actual behavior and has found that they often lack faithfulness.
However, the question of how to improve faithfulness remains underexplored.
Moreover, because different explanation styles have superficially distinct characteristics, it is unclear whether improvements observed in one style also arise when using other styles.
This study analyzes the effects of training for faithful self-explanations and the extent to which these effects generalize, using three classification tasks and three explanation styles.
We construct one-word constrained explanations that are likely to be faithful using a feature attribution method, and use these pseudo-faithful self-explanations for continual learning on instruction-tuned models.
Our experiments demonstrate that training can improve self-explanation faithfulness across all classification tasks and explanation styles, and that these improvements also show signs of generalization to the multi-word settings and to unseen tasks.
Furthermore, we find consistent cross-style generalization among three styles, suggesting that training may contribute to a broader improvement in faithful self-explanation ability.
\end{abstract}

\section{Introduction}
Instruction-tuned large language models (LLMs) appear capable of generating natural language explanations about their own decisions (i.e., self-explanations) in a variety of styles~\cite{calderon-2025-stakeholders}.
Users can instruct the models to identify the key information in the input that drives their predictions or to construct counterfactual inputs that invert the original predictions.
Self-explanations have the potential to give explainability to LLMs, converting their black-box processing into interpretable expressions.
\begin{table}[t]
\centering
\resizebox{0.8\columnwidth}{!}{%
\begin{tabular}{|ll|}
\hline
\rowcolor{gray!20}\multicolumn{2}{|c|}{\textbf{Train - Attribution Style}} \\ \hline%
User: & \begin{tabular}[c]{@{}l@{}}Text: ``I hate waking up early.''\\ What is the sentiment of the text?\end{tabular} \\ \hline
Assistant: & Negative \\ \hline
User: & \begin{tabular}[c]{@{}l@{}}List the most important word for\\ determining the sentiment.\end{tabular} \\ \hline
Assistant: & \textbf{``hate''} \\ \hline\noalign{\vspace{4pt}}\hline
\rowcolor{gray!20}\multicolumn{2}{|c|}{\textbf{Test - Counterfactual Style}} \\ \hline%
User: & \begin{tabular}[c]{@{}l@{}}Text: ``my room walls are boring''\\ What is the sentiment of the text?\end{tabular} \\ \hline
Assistant: & Negative \\ \hline
User: & \begin{tabular}[c]{@{}l@{}}Edit the text so that the predicted \\ sentiment would change.\end{tabular} \\ \hline
Assistant: & \textbf{``my room walls are exciting''} \\ \hline
\end{tabular}%
}
\caption{Training and test examples used to evaluate cross-style generalization. The training style instructs models to output \textit{words} that strongly \textit{support} their predictions, whereas the test style requires generating \textit{sentences} that \textit{contradicts} the predictions.}
\label{tab:overview}
\end{table}

Recent studies have investigated the extent to which self-explanations faithfully reflect actual model behavior.
They have designed evaluation protocols for each explanation style: checking the model's prediction change when editing the input according to the self-explanations~\cite{atanasova-etal-2023-faithfulness, siegel-etal-2024-probabilities, madsen-etal-2024-self}, and checking whether the Chain-of-Thought reasoning steps accurately reflect their biases in predictions~\cite{turpin-etal-2023-what, chen-2025-reasoningmodels-dontsay}.
These studies show that self-explanations produced by LLMs are often unfaithful and unreliable across styles, underscoring the need for improvement.

\begin{figure*}[t]
    \centering
    \includegraphics[width=\textwidth]{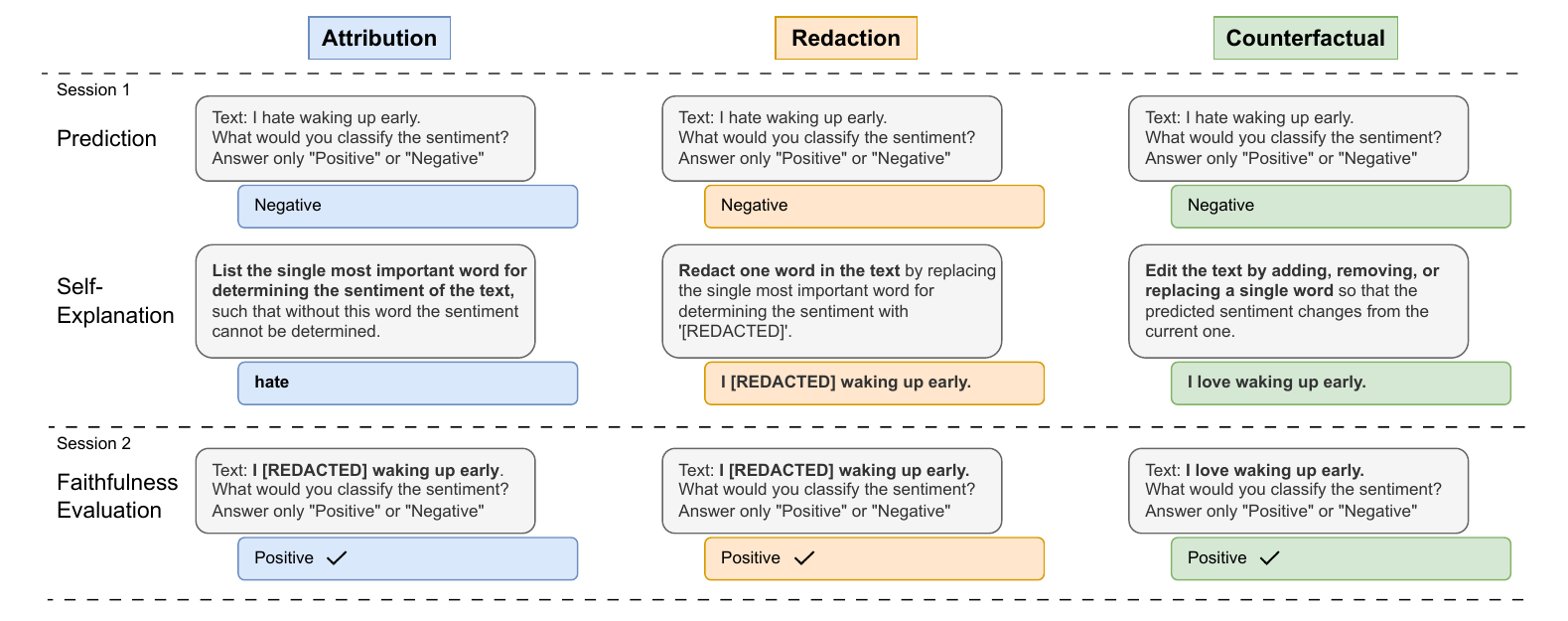}
    \caption{Examples of one-word-constrained self-explanations and faithfulness evaluation for each explanation style. Self-explanations are generated in the same session as the classification task: Attribution and Redaction require listing and redacting the most important input words affecting the prediction, respectively, while Counterfactual requires editing the input text so that the predicted label will flip. Faithfulness evaluation involves a separate session, in which a self-explanation is considered faithful if editing the input according to it indeed flips the prediction.}
    \label{fig:faithfulness_test}
\end{figure*}

However, it remains unclear how, and to what extent, the self-explanation faithfulness of LLMs can be improved.
Because the faithfulness of a model’s self-explanation should be evaluated based on observations of its own behavior~\cite{jacovi-goldberg-2020-towards}, it is inherently challenging to provide general supervised signals of faithful self-explanations that can apply to any model.
Moreover, explanation styles exhibit distinct surface characteristics: an attribution-style self-explanation consists of words that support the model’s prediction, whereas a counterfactual-style explanation is expressed through a sentence that contradicts the original prediction (Table~\ref{tab:overview}).
It remains an open question how the faithfulness of self-explanations in each explanation style can be effectively improved, and whether such improvements are transferable across styles.

In this paper, we construct pseudo-faithful self-explanations in three explanation styles (Figure~\ref{fig:faithfulness_test}) and examine how training LLMs on these constructed explanations affects their faithfulness.
We further investigate how well the resulting improvements generalize along three dimensions: unconstrained multi-word settings (Section~\ref{subsec: Generalization to Multi-Word Explanation Styles}), unseen classification tasks (Section~\ref{subsec: Generalization Across Classification Tasks}), and cross-style generalization (Section~\ref{subsec: Generalization Across Faithfulness Tests}).
We construct training datasets of pseudo-faithful self-explanations for three classification tasks using a feature attribution method under a one-word constrained setting.
We then train the instruction-tuned models by mixing the constructed self-explanations with their original instruction tuning data, and evaluate the self-explanation faithfulness before and after training.

Our experimental results show that training improves faithfulness across almost all classification tasks and explanation styles.
We also find that, for one explanation style, the improvement generalizes to unseen classification tasks and to unconstrained multi-word settings.
Furthermore, we observe generalization of faithfulness improvements across distinct explanation styles.
For example, a model trained to identify words that support its prediction can also modify the input sentence by deleting or replacing those words to invert the prediction.
These findings suggest that training on pseudo-faithful self-explanations may improve self-explanation faithfulness across explanation styles, even without access to truly faithful self-explanations.

\section{Explanation Styles and Faithfulness}
\label{sec:Self-Explanation Styles and Faithfulness}
Previous work has proposed a variety of explanation styles and corresponding protocols for assessing their faithfulness to model behavior.
A common style requires models to produce self-explanations consisting of input words identified as contributing to predictions~\cite{atanasova-etal-2023-faithfulness,huang-20230largelanguagemodel-selfexplain,madsen-etal-2024-self}, while more free-form explanations have also been explored~\cite{siegel-etal-2024-probabilities}.
Another line of research adopts a counterfactual style, in which explanations take the form of sentences similar to the original input but intended to induce different predictions~\cite{singh2024-rethinkinginterpretability-eralarge,calderon-2025-stakeholders}.  
In this setting, faithfulness can be evaluated by checking whether the generated counterfactuals indeed produce the prediction change.

We focus on three styles of self-explanations, namely attribution, redaction, and counterfactual, and evaluate their faithfulness primarily through the self-consistency check protocol~\cite{madsen-etal-2024-self}, as illustrated in Figure~\ref{fig:faithfulness_test}.
We describe the details as follows:

\paragraph{Attribution}
In this style, the model lists input words that it considers important for its prediction, thereby simulating feature attribution methods.
If the explanation is faithful, the listed words should have a substantial impact on the prediction being explained.
Faithfulness is therefore assessed by examining whether the prediction changes when the listed words are removed from the original input.
Following \citet{madsen-etal-2024-self}, we create such redacted inputs by automatically replacing the listed input words with the ``[REDACTED]'' tokens rather than deleting them, in order to preserve the grammatical structure.

\paragraph{Redaction}
In this style, the model directly generates a redacted version of the input in which the words it deems important for its prediction are replaced with ``[REDACTED]''.
Unlike attribution, which requires the model to list important words, the redaction style requires the model to erase them while preserving the rest of the input sentence.
We evaluate faithfulness by checking whether the model’s prediction changes when it is given the redacted input sentence it produced.

\paragraph{Counterfactual}
This style requires the model to edit the input sentence such that the resulting sentence changes the model’s original prediction.
The model may add, remove, or replace input words, subject to editing-distance constraints specified in a prompt.
To evaluate faithfulness, we feed the generated counterfactual sentences back into the model and test whether the predicted label changes accordingly.

It is important to note that these explanation styles differ substantially in their surface forms: whether a self-explanation is a sentence or a list of words, whether it involves adding new content beyond the original input, and whether it supports or contradicts the original prediction.

\section{Training for Faithful Self-Explanations}
\label{sec: Self-Explanation Training}
Our goal is to analyze how training models with faithful self-explanations improves faithfulness and how these improvements generalize.
We do not have access to the ground truth of truly faithful self-explanations as a principle~\cite{jacovi-goldberg-2020-towards}; faithfulness is defined through the model's black-box behavior and evaluated by checking the consistency of generated self-explanations in a post-hoc manner.
We therefore consider pseudo-faithful self-explanations that are more likely to be judged as faithful, rather than attempting to construct genuinely faithful ones.
We first create datasets of pseudo-faithful self-explanations for each of the three styles, using influential words estimated via a feature attribution method.
We then train models on these datasets in a continual learning setup and evaluate the effects using the faithfulness evaluation protocols for each style.

\subsection{Training Dataset Construction}
For all of our experiments, we construct training datasets of pseudo-faithful self-explanations using instruction-tuned Llama-2~\cite{touvron-2023-llama2} models, specifically Tulu-2~\cite{ivison-2023-tulu2} 7B and 13B, and three classification tasks: Sentiment140~\cite{go-2009-twitter}, SNLI~\cite{bowman-etal-2015-snli}, and AGNews\footnote{\url{https://www.kaggle.com/datasets/amananandrai/ag-news-classification-dataset}}.
We assume that faithful explanations, including attribution, redaction, and counterfactual styles, are generally expected to capture the causal influence of input words on model predictions.
For this reason, we hypothesize that pseudo-faithful self-explanations can be constructed from the most influential input word identified by a feature attribution method.

\paragraph{Influential Word Estimation}
The influence of each input word is estimated using an erasure-based attribution method~\cite{li-2017-erasure}.
Let the input sentence be $x = (w_1, w_2, \ldots, w_m)$ and the model prediction be $\hat{y} = \mathop{\arg\max}_{y} p_\theta(y \mid x)$, where $\theta$ denotes the model.
We compute the influence of an input word $w$ on the prediction $\hat{y}$:
\begin{equation}
    I_\theta(w \mid x)=p_\theta(\hat{y} \mid x) - p_\theta(\hat{y} \mid x_{-w}),
\end{equation}
where $x_{-w}$ is obtained by replacing $w$ with the ``[REDACTED]'' token.
We then identify the word $w^*$ with the highest value as the most influential word on their prediction:
\begin{equation}
\label{eqn:ground_truth_explanation}
    w^* = \mathop{\arg\max}_{w \in x} I_\theta(w \mid x).
\end{equation}

\begin{table}[t]
\resizebox{\columnwidth}{!}{%
\begin{tabular}{|ll|}
\hline
User: & \begin{tabular}[c]{@{}l@{}}Text: \{input $x$\}\\ What is the sentiment of text?\end{tabular} \\ \hline
Assistant: & \{model prediction $\hat{y}$\} \\ \hline
User: & \{self-explanation instruction for style $S$\} \\ \hline
\textbf{Assistant:} & \textbf{\{constructed self-explanation\}} \\ \hline
\end{tabular}%
}
\caption{Template of the training data. The loss is computed solely from \textbf{the responses of self-explanations}.}
\label{tab:training_data_template}
\end{table}

\paragraph{Construction of Pseudo-Ground Truth}
Using the identified influential word $w^*$, we construct pseudo-ground truth of faithful self-explanations for each style.
For all styles, we constrain the self-explanations to a one-word setting~(Figure~\ref{fig:faithfulness_test}).
The construction procedure is as follows:
\begin{itemize}
    \item \textbf{Attribution}: 
    The pseudo-ground truth self-explanation is simply the influential word $w^*$ corresponding to the model’s prediction $\hat{y}$ (e.g., \texttt{hate}).
    
    \item \textbf{Redaction}:
    The pseudo-ground truth self-explanation is the redacted input $x_{-w^*}$, created by replacing $w^*$ with “[REDACTED]” (e.g., \texttt{I [REDACTED] waking up early.}).
    
    \item \textbf{Counterfactual}:
    The pseudo-ground truth self-explanation is constructed by replacing $w^*$ with another word $w_{\bar{y}}$ associated with the second most probable prediction $\bar{y}$ (e.g., \texttt{I love waking up early.}).
    We obtain $w_{\bar{y}}$ by prompting the Tulu-2 models with the following instruction:
\end{itemize}
\begin{tcolorbox}[
        colback=white,
    ]
        \texttt{{Redacted sentence: \{$x_{-w^*}$\}\\Replace ``[REDACTED]'' with exactly one word that would make the completed sentence very likely to be predicted with the \{${\bar{y}}$\}}. \\Output word:}
\end{tcolorbox}
We then convert the pseudo-ground truth self-explanations for each style into training examples using a template exemplified in Table~\ref{tab:training_data_template}.
Self-explanation instructions follow the format shown in Figure~\ref{fig:faithfulness_test}, with additional details provided in Appendix~\ref{sec:appendix_prompt}.

Our dataset construction procedure aims to generate pseudo-ground truth self-explanations that are more faithful than originally produced self-explanations, rather than attempting to obtain fully faithful explanations, which are unavailable.
As shown in Table~\ref{tab:training data validation}, we validate the quality of our constructed datasets by ensuring that the faithfulness scores (Section~\ref{subsec:evaluation}) of the training samples exceed those of the originally generated ones\footnote{The constructed self-explanations for the attribution and redaction styles are expected to yield the same faithfulness scores, as they are evaluated using the same redacted inputs.}.

\begin{table}[t]
\centering
\resizebox{0.9\columnwidth}{!}{%
\begin{tabular}{llccc}
\hline
 &  & Attribution & Redaction & Counterfact \\
 &  & One-word & One-word & One-word \\ \hline
\footnotesize{Tulu-2 7B} &  &  &  &  \\
Original &  & 0.124 & 0.124 & 0.186 \\
Constructed &  & \textbf{0.342} & \textbf{0.342} & \textbf{0.331} \\ \hline
\footnotesize{Tulu-2 13B} &  &  &  &  \\
Original &  & 0.134 & 0.090 & 0.335 \\
Constructed &  & \textbf{0.304} & \textbf{0.304} & \textbf{0.435} \\ \hline
\end{tabular}%
}
\caption{Comparison of faithfulness scores between self-explanations originally generated by the models and constructed ones included in the training dataset, each evaluated on 1,000 samples from Sentiment140.}
\label{tab:training data validation}
\end{table}

\begin{table*}[t]
\centering
\resizebox{0.9\textwidth}{!}{%
\begin{tabular}{llccclccclccc}
\hline
\multicolumn{1}{c}{} &  & \multicolumn{3}{c}{Attribution} &  & \multicolumn{3}{c}{Redaction} &  & \multicolumn{3}{c}{Counterfactual} \\
 &  & \multicolumn{3}{c}{One-word} &  & \multicolumn{3}{c}{One-word} &  & \multicolumn{3}{c}{One-word} \\ \cline{3-5} \cline{7-9} \cline{11-13} 
 &  & Sent140 & SNLI & AGNews &  & Sent140 & SNLI & AGNews &  & Sent140 & SNLI & AGNews \\ \hline
\footnotesize{Tulu-2 7B} &  &  &  &  &  &  &  &  &  &  &  &  \\
No-Training &  & 0.120 & 0.199 & \textbf{0.248} &  & 0.102 & 0.244 & 0.237 &  & 0.173 & 0.076 & 0.087 \\
w/ Predictions &  & 0.126 & 0.161 & 0.127 &  & 0.099 & 0.282 & 0.136 &  & 0.129 & 0.079 & 0.035 \\
w/ Explanations &  & \textbf{0.300} & \textbf{0.457} & 0.199 &  & \textbf{0.271} & \textbf{0.355} & \textbf{0.323} &  & \textbf{0.241} & \textbf{0.170} & \textbf{0.249} \\ \hline
\footnotesize{Tulu-2 13B} &  &  &  &  &  &  &  &  &  &  &  &  \\
No-Training &  & 0.140 & 0.177 & 0.185 &  & 0.110 & 0.317 & 0.149 &  & 0.303 & \textbf{0.243} & 0.049 \\
w/ Predictions &  & 0.141 & 0.182 & 0.099 &  & 0.080 & \textbf{0.335} & 0.077 &  & 0.270 & 0.216 & 0.027 \\
w/ Explanations &  & \textbf{0.255} & \textbf{0.299} & \textbf{0.281} &  & \textbf{0.204} & 0.306 & \textbf{0.265} &  & \textbf{0.595} & 0.192 & \textbf{0.417} \\ \hline
\end{tabular}%
}
\caption{Faithfulness scores, measured as the proportion of faithful self-explanations (Section~\ref{subsec:evaluation}) before and after training. ``No-Training'' refers to the off-the-shelf model before training, ``w/~Predictions'' refers to models trained with ground-truth predictions for the classification tasks, and ``w/~Explanations'' refers to models trained with the constructed pseudo-faithful self-explanations for each style conditioned on their own predictions.}
\label{tab:training effect}
\end{table*}

\subsection{Continual Learning}
We train the Tulu-2 7B and 13B models using the constructed self-explanation datasets in a continual learning setting.
Preventing catastrophic forgetting~\cite{luo-2023-catastrophic} is particularly important in our experiments, as the faithfulness evaluation and our analysis of generalization require the models to maintain performance on multiple tasks beyond the training setting.
To mitigate forgetting, we mix the instruction-tuning data originally used for training the Tulu-2 models during continual learning~\cite{scialom-etal-2022-fine}.
We apply Low-Rank Adaptation~\cite[LoRA;][]{hu-2021-lora}, training for one epoch with 50{,}000 samples from the constructed self-explanation dataset and 10{,}000 samples from the instruction-tuning data.

\subsection{Evaluation}
\label{subsec:evaluation}
We evaluate the faithfulness of the models’ self-explanations before and after training as the proportion of self-explanations judged faithful using the self-consistency check (Section~\ref{sec:Self-Explanation Styles and Faithfulness}).
Specifically, we first collect the model’s predictions on 5,000 samples that do not overlap with the training data, together with self-explanations for each style.
For each style, we then edit the inputs according to the generated self-explanations and compute faithfulness as the proportion of cases in which the model’s prediction changes.

We exclude instances that violate either the style condition or the number-of-word condition\footnote{The number of evaluation instances retained for each experiment is reported in Table~\ref{tab:number_of_evaluated_instances}}.
The style condition requires that self-explanations:
(i) list only the input words in the attribution style,
(ii) include the ``[REDACTED]'' tokens without altering the remaining input in the redaction style, and
(iii) edit the input without using ``[REDACTED]'' tokens or the classification label itself (e.g., ``Positive'') in the counterfactual style.
Because the prompts explicitly instructed the models to satisfy these requirements, violations indicate failures in instruction following rather than evidence of unfaithfulness.
The number-of-word condition retains only the self-explanations in which the model lists $N$ words in the attribution style, redacts $N$ words in the redaction style, and modifies the input with an edit distance of $N$ in the counterfactual style ($N = 1, 2, 3, 4, 5$).
This condition ensures fair comparison; for example, if a model lists, redacts, or edits an excessively large number of words in its self-explanation, it may be judged faithful in an unfair manner.
We set $N=1$ in most experiments, instructing the model to produce one-word constrained self-explanations for each style to match the training setup.
In Section~\ref{subsec: Generalization to Multi-Word Explanation Styles}, we also evaluate faithfulness for $N=2,3,4,5$ in a generalized multi-word setting, using prompts that instruct the model to list, redact, or edit any number of input words for each style.

\section{Results}
\label{sec: Results}

\subsection{Training Effects}
\label{subsec: Effects of Self-Explanation Training}
We first examine the interpolation effects of training by evaluating the faithfulness of the models before and after training under the same settings used during training.
In addition to the off-the-shelf models, we include a baseline in which models are trained using the ground-truth predictions for the classification tasks.

Table~\ref{tab:training effect} shows that models trained with the constructed self-explanation datasets produce more faithful self-explanations than the off-the-shelf models in most settings.
For example, the 13B models trained with self-explanations improve by 0.115, 0.094, and 0.292 points in the attribution, redaction, and counterfactual styles, respectively, on the sentiment analysis task (Sentiment140).
These results empirically confirm that training with pseudo-faithful self-explanations can enhance self-explanation faithfulness, even without access to true ``ground-truth'' faithful explanations.

By contrast, models trained with the ground-truth predictions for the classification tasks often show improvements of less than 0.01 or even a decrease in self-explanation faithfulness.
This demonstrates that faithfulness is improved specifically by training on the constructed self-explanations conditioned on the models' own predictions, rather than by training on ground-truth predictions for the classification tasks\footnote{The performance on the classification task is reported in Appendix~\ref{subsec: Classification Task Performance}, and is not significantly changed after training.}.

\begin{table}[t]
\resizebox{0.97\columnwidth}{!}{%
\begin{tabular}{lccc}
\hline
\multicolumn{1}{c}{\multirow{2}{*}{}} & Attribution & Redaction & Counterfactual \\
\multicolumn{1}{c}{} & Multi-word & Multi-word & Multi-word \\ \hline
\footnotesize{Tulu-2 7B} &  &  &  \\
No Training & 0.216 & 0.074 & \textbf{0.246} \\
w/ Explanations & \textbf{0.451} & \textbf{0.154} & 0.231 \\ \hline
\footnotesize{Tulu-2 13B} &  &  &  \\
No Training & 0.234 & 0.125 & 0.345 \\
w/ Explanations & \textbf{0.435} & \textbf{0.174} & \textbf{0.497} \\ \hline
\end{tabular}%
}
\caption{Faithfulness scores for the Sentiment140 dataset in the unconstrained multi-word setting. “w/~Explanations” models are trained using one-word constrained self-explanations for each style.}
\label{tab:multi-word_faithfulness}
\end{table}

\begin{figure*}
    \centering
    \includegraphics[width=0.9\textwidth]{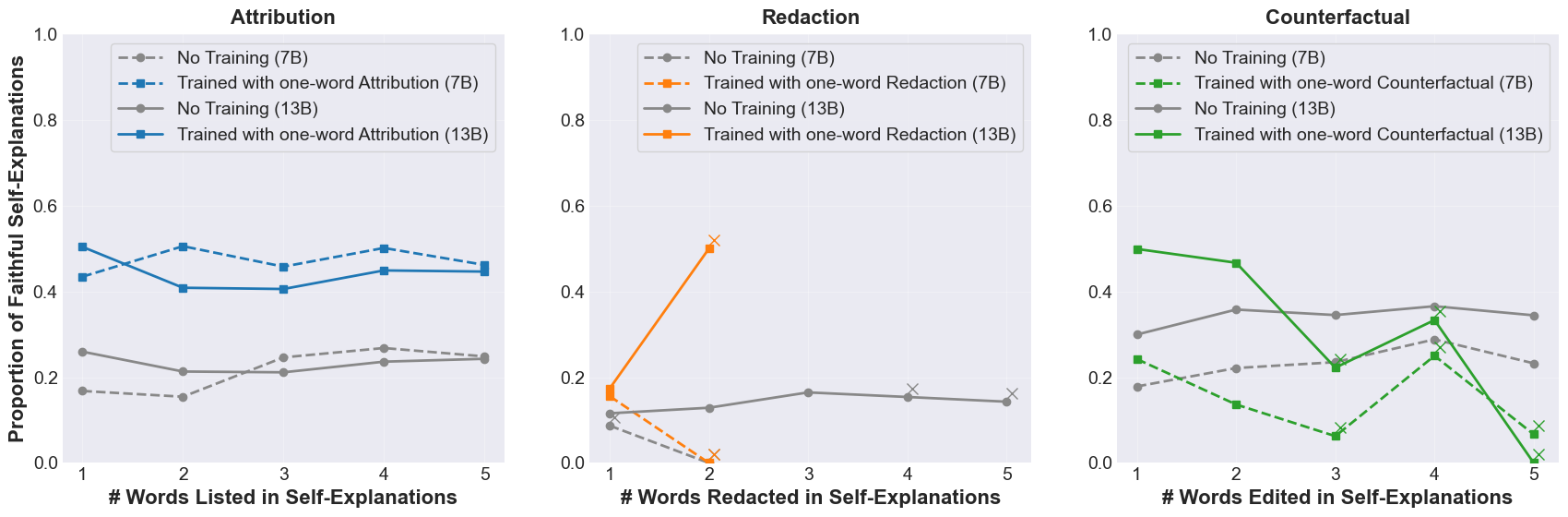}
    \caption{Evaluation of the generalization to the multi-word setting (Section~\ref{subsec: Generalization to Multi-Word Explanation Styles}) on the Sentiment140 dataset. We report the proportion of faithful self-explanations for each number of words that are used in the self-explanations for each style. Data plots marked with ``$\times$'' indicate that the number of evaluation instances is less than 50.}
    \label{fig:gen_number_of_words}
\end{figure*}

\begin{figure*}[t]
    \centering
    \includegraphics[width=0.9\textwidth]{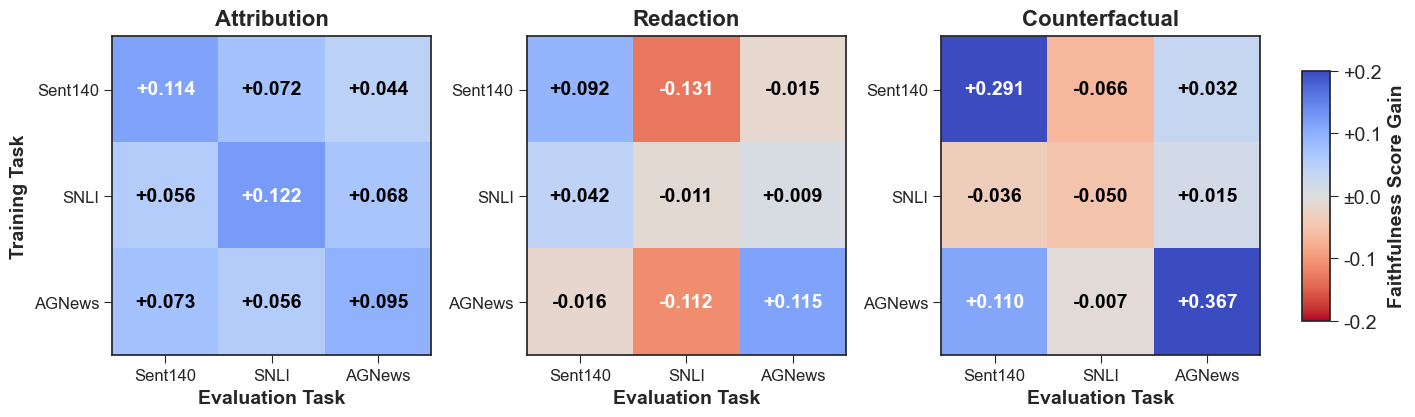}
    \caption{Evaluation of the generalization across different classification tasks. For each training-evaluation task pair, we measure the faithfulness score gain before and after training with self-explanations, defined as the increase or decrease in the proportion of faithful self-explanations. Results are reported using the Tulu-2 13B model.}
    \label{fig:gen_classification_task}
\end{figure*}

\subsection{Generalization to Multi-Word Setting}
\label{subsec: Generalization to Multi-Word Explanation Styles}
During training, the models learn to generate self-explanations in the one-word setting, where they are permitted to list, redact, or edit only a single input word.
However, self-explanations in practice are not necessarily restricted to a single word, since interactions among multiple words may be required to express certain meanings.
We therefore introduce a multi-word setting using prompts that permit the model to use any number of words in its self-explanations, rather than enforcing a one-word constraint, as illustrated below:
\begin{tcolorbox}[
    colback=white,
]
\texttt{List \textbf{all and only} the most important words for determining the sentiment.}
\end{tcolorbox}
\noindent We focus on the Sentiment140 dataset because the trained models consistently exhibit improvements across all three styles on this dataset.

We first measure faithfulness as the proportion of self-explanations that are judged as faithful (Section~\ref{sec:Self-Explanation Styles and Faithfulness}) and satisfy the style condition, while removing the number-of-word condition (Section~\ref{subsec:evaluation}).
As shown in Table~\ref{tab:multi-word_faithfulness}, the models trained with one-word self-explanations achieve higher faithfulness scores even when multi-word self-explanations are allowed.
This suggests that their advantage is maintained beyond the one-word setting.

We further examine whether improvements in faithfulness occur for each word count that the model lists, redacts, or edits in its explanations ($N=1,2,3,4,5$).
Specifically, we group self-explanations by the number of words listed, redacted, or edited for each style, and compute the proportion of faithful self-explanations within each group.
Figure~\ref{fig:gen_number_of_words} shows that, only in the attribution style, models trained with one-word self-explanations consistently generate more faithful explanations across different numbers of used words.
These findings indicate that generalization to multi-word settings depends on the style and may emerge exclusively in the attribution style.

\begin{figure*}
    \centering
    \includegraphics[width=\textwidth]{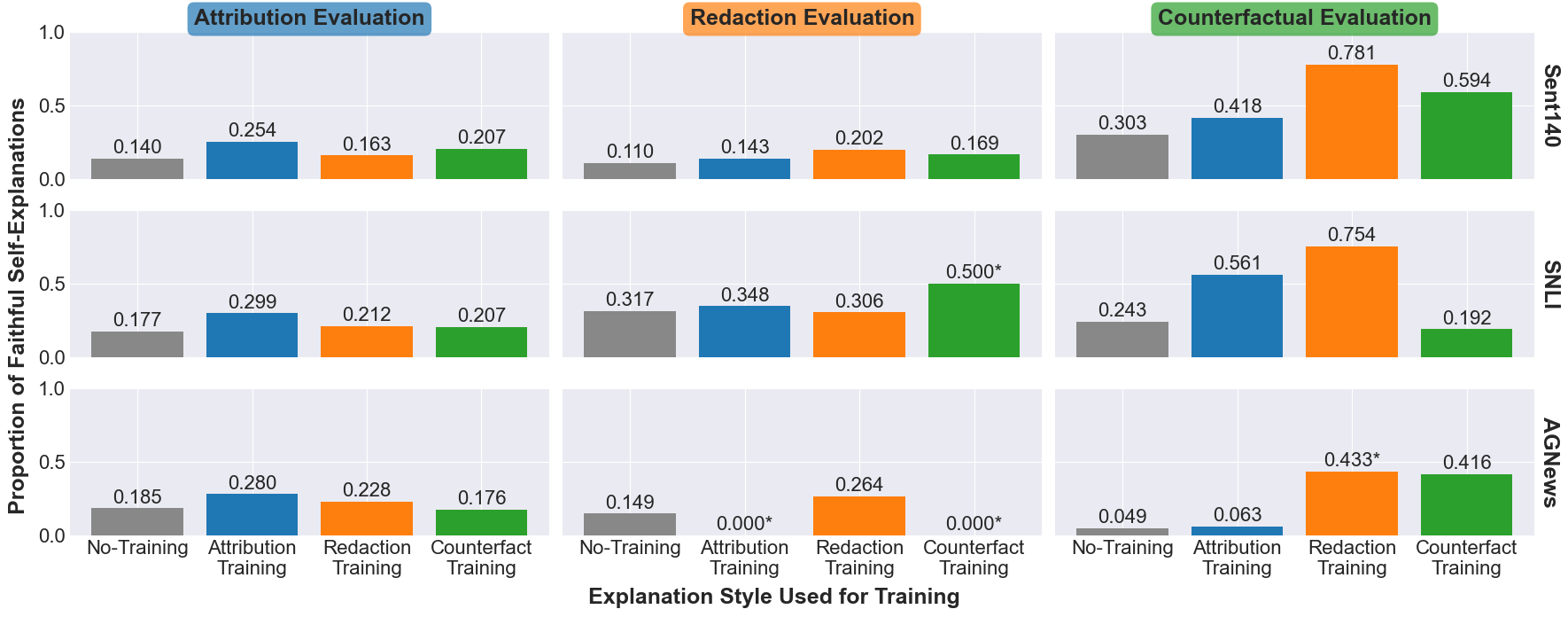}
    \caption{Evaluation of generalization across explanation styles. Each value represents the faithfulness score obtained under a given evaluation style, and each training condition specifies the explanation style used for training. Values marked with ``*'' indicate that the number of evaluation instances is less than 50. Results are reported using the Tulu-2 13B model.}
    \label{fig:gen_styles}
\end{figure*}

\subsection{Generalization across Classification Tasks}
\label{subsec: Generalization Across Classification Tasks}
We have observed that, for a given classification task, training improves the faithfulness of the model’s self-explanations for each style.
A natural question is whether such training also improves faithfulness on unseen classification tasks.

Figure~\ref{fig:gen_classification_task} reports the gains in faithfulness scores relative to the off-the-shelf models, evaluated across different combinations of training and evaluation tasks.
We find consistent faithfulness improvements in the attribution style: for example, models trained with attribution-style explanations on the Sentiment140 dataset achieve increases of 0.072 points on SNLI and 0.044 points on AGNews.
In contrast, under the redaction and counterfactual styles, the models struggle to generate faithful self-explanations for unseen classification tasks.
These results indicate that, mirroring the trend observed in the multi-word setting, whether the training effects generalize across classification tasks depends on the explanation style; generalization is most reliably observed in the attribution style.

\subsection{Generalization across Explanation Styles}
\label{subsec: Generalization Across Faithfulness Tests}
We have examined the effects of training and its generalization within each explanation style.
We next investigate whether training generalizes across explanation styles.
Such cross-style generalization is practically important, as real-world explanation styles are often more diverse and more free-form than those included in our experiments.

We evaluate self-explanation faithfulness using styles that the models did not encounter during training.
As before, faithfulness is measured as the proportion of self-explanations that are judged faithful and that satisfy the conditions; for instance, if a model trained on the redaction style produces a self-explanation containing the “[REDACTED]” token in the counterfactual style, that instance is excluded because it violates the prompt instructions.

Figure~\ref{fig:gen_styles} shows the proportion of faithful self-explanations for each training–evaluation style pair, comparing the results before and after training.
We observe improvements in faithfulness even when the training and evaluation styles differ.
For instance, on the Sentiment140 dataset (top row), models trained using attribution-style explanations (blue bars) generate more faithful self-explanations than the untrained models (black bars) even when evaluated using the redaction or counterfactual styles, which were unseen during training.
These improvements are notable given that the attribution style requires the model to output input words that support their predictions, whereas the redaction and counterfactual styles require the model to generate sentences that contradict them.
This suggests that the training effects can transfer across different styles, rather than being confined to the style used during training.

\begin{table*}[t]
\centering
\resizebox{0.68\textwidth}{!}{%
\begin{tabular}{lll}
\hline
\textbf{\begin{tabular}[c]{@{}l@{}}Explanation\\ Style\end{tabular}} & \textbf{\begin{tabular}[c]{@{}l@{}}Classification\\ Task\end{tabular}} & \textbf{\begin{tabular}[c]{@{}l@{}}Top-10 Frequent Words in\\ Faithful Self-Explanations\end{tabular}} \\ \hline
Attribution & Sentiment140 & \begin{tabular}[c]{@{}l@{}}not, no, good, don't, miss, hate, sad, \\ can't, love, bad\end{tabular} \\ \hline
Attribution & AGNews & \begin{tabular}[c]{@{}l@{}}Iraq, Afghan, Arafat, Iran, Oracle, Putin,\\ Google, Baghdad, Microsoft, Stocks\end{tabular} \\ \hline
Counterfactual & Sentiment140 & \begin{tabular}[c]{@{}l@{}}DELETION*, happy, hate, good, bad, \\ love, terrible, worse, great\end{tabular} \\ \hline
\end{tabular}%
}
\caption{Examples of the most frequent words appearing in faithful self-explanations for each setting, generated by the Tulu-2 13B model trained with attribution-style self-explanations on the Sentiment140 dataset. For the counterfactual style, the listed words correspond to words replaced or added relative to the original input, and ``DELETION*'' indicates that a certain word is removed from the input.}
\label{tab:exampled_of_words_used_in_self-explanations}
\end{table*}

\section{Discussion}
\label{sec:Discussion}
We observe that the improvements from training can generalize across classification tasks and across explanation styles.
However, one might suspect that models simply acquire heuristics tailored to the evaluation protocol and therefore behave consistently across different evaluation settings.
Although a truly faithful self-explanation cannot be predefined in principle, trained models are not expected to produce self-explanations in a uniform manner across conditions, even when these explanations are judged as faithful.
This raises a question: do the trained models rely on fixed heuristics regardless of the setting, or do they acquire a more general capability for generating faithful explanations across different conditions?

To answer this question, we qualitatively analyze the generated self-explanations that are judged as faithful during evaluation.
Table~\ref{tab:exampled_of_words_used_in_self-explanations} reports the lemmatized words generated in self-explanations from the Tulu-2 13B model trained with attribution-style explanations on the sentiment analysis task (Sentiment140).
In the training setting of attribution-style explanations for sentiment analysis, the model tends to generate negation expressions (e.g., ``no'', ``can't''), as well as words associated with emotions (e.g., ``hate'', ``love'').
In self-explanations for the unseen topic classification task (AGNews), however, the same model generates different types of words, including proper nouns (e.g., ``Iraq'', ``Google'') and business words (e.g., ``Stocks'').
We also observe such vocabulary differences across explanation styles.
In unseen counterfactual-style explanations for sentiment analysis, the model frequently produces sentiment-bearing words (e.g., ``hate'', ``terrible'') as expected; however, it does not use negation expressions, which are common in the attribution-style setting used during training.
These observations may suggest that the models after training could generate faithful self-explanations generally to the given classification tasks and styles, rather than depending on fixed heuristics tailored to the evaluation protocol of the training style.

\section{Related Work}
Researchers have investigated how faithfully the intermediate reasoning chains generated by LLMs reflect their final decisions under Chain-of-Thought prompting~\cite{turpin-etal-2023-what, chen-2025-reasoningmodels-dontsay}.
In evaluations of CoT faithfulness, prior work introduces typical forms of bias that alter the model's prediction, such as inserting phrases like ``I think the answer is (A),'' and shows that the resulting CoT reasoning steps often fail to reflect these inserted biases~\cite{turpin-etal-2023-what, matton-2024-walk}.
Recent studies have suggested that reasoning models, which are trained via reinforcement learning to improve general CoT performance, exhibit higher CoT faithfulness than non-reasoning models, though there remains room for improvement.~\cite{chen-2025-reasoningmodels-dontsay,chua-2025-deepseekr1-reasoningmodels}.

Our study focuses on three explanation styles other than CoT and examines both the training effects and their generalization when using supervised signals explicitly designed to promote faithful self-explanations.
It is worth noting that constructing pseudo-faithful CoT reasoning steps is inherently difficult, because each intermediate reasoning step must influence subsequent steps as well as the final prediction.

\section{Conclusion}
We investigated how training affects the faithfulness of LLM self-explanations and the extent to which these effects generalize.
To address the lack of access to truly faithful explanations, we constructed pseudo-ground truth data of faithful self-explanations under a one-word constrained setting using an attribution method.
Our experiments demonstrated that training generally improves self-explanation faithfulness across classification tasks and explanation styles.
We further found evidence that these improvements can generalize to the unconstrained multi-word setting and to unseen classification tasks.
In addition, we observed consistent cross-style generalization, indicating that the benefits of training extend beyond individual explanation styles.
We believe that our findings on faithfulness contribute to advancing the understanding and improvement of LLM trustworthiness.

\section*{Limitations}
The training procedure in our experiments requires access to the trained model's instruction-tuning data.
This requirement limits the applicability of similar investigations to models for which such training data is publicly available.
Although we incorporate multiple classification tasks commonly used in the faithfulness evaluation literature, the scope of tasks remains limited, excluding more complex settings such as generative tasks.
Moreover, our training and evaluation primarily focus on simple explanations involving single-word operations, with existing but only limited assessment of generalization to more complex, freer-format setups.
Finally, as our primary scope is the evaluation of self-explanation faithfulness, we leave other evaluation perspectives for future work, particularly examining whether the observed improvements contribute to human-centered explainability, such as simulatability~\cite{hase-bansal-2020-simulatability}.

\section*{Ethics Statement}
Although our procedures for constructing the self-explanation dataset do not involve any explicit gender bias or abusive language, there remains the possibility that such biases could be inherited from the models or datasets used in our experiments.
We caution that users of LLMs should not place unwarranted trust in a model’s self-explanations without careful consideration, regardless of whether the model was trained following our procedures.
We hope that this work will contribute to future research aimed at analyzing and enhancing the trustworthiness of LLMs, thereby supporting sound and responsible human decision-making.

\section*{Acknowledgements}
We thank the three anonymous reviewers for their helpful comments and feedback.
This work was partially supported by JSPS KAKENHI Grant Number JP24H00809, JST BOOST Grant Number JPMJBY24H5, and JST SPRING Grant Number JPMJSP2108.

\bibliography{custom}

\newpage
\appendix

\section{Dataset}
\label{sec:appendix_dataset}
We employ three classification datasets in Table~\ref{tab:exampled_of_classification_datasets}: Sentiment140 for the binary sentiment analysis, SNLI for the ternary NLI task, and AGNews for the quaternary topic classification.
For our experiments, we sample almost 50,000 examples for training and nearly 5,000 samples for evaluation from each dataset, ensuring that the class labels are balanced.
The statistics of these examples are shown in Table~\ref{tab:classification_dataset_statistics}.

\section{Prompt}
\label{sec:appendix_prompt}
We present prompt templates for classification and self-explanation tasks on Sentiment140 in Table~\ref{tab:prompts_for_sentiment140}, and those for SNLI and AGNews in Table~\ref{tab:prompts_for_snli_agnews}.
Although all prompt designs largely follow those introduced by~\citet{madsen-etal-2024-self}, we include additional instructions for the response format in the self-explanation tasks, such as ``one word following Answer:'' and ``answer in JSON format.''
The Tulu-2 models sufficiently adhere to these format instructions in the experiments, enabling a fair evaluation of their performance in the self-explanation task without major formatting issues.

In Table~\ref{tab:prompts_counterfactual_construction}, we show the prompts used for obtaining the word $w_{\bar{y}}$, which is expected to be associated with the second probable prediction $\bar{y}$, to construct the counterfactual self-explanation datasets.
The instruction includes prohibiting the use of the prediction label itself or the ``[REDACTED]'' token, to prevent a skeptical shortcut for the counterfactual self-explanations.
We also automatically filter out such instances to ensure exclusion.

\section{Hyperparamters}
\label{sec:appendix_hyperparameters}
For text generation, the temperature is set to 0, and the number of beam searches is 1, enabling the Tulu-2 models to generate tokens one by one in a deterministic greedy manner.
This setting ensures reproducibility without any randomness; we conduct the experiments only once.
For continual learning, we mainly adopt the setting used for instruction tuning with LoRA in the Tulu-2 models.
Specifically, the learning rate is set to 1e-4, the LoRA rank is set to 64, the value of $\alpha$ is set to 16, and the dropout rate is set to 0.1.
All attention layers are designated as trainable modules, and the model is trained for one epoch.

\section{Classification Task Performance}
\label{subsec: Classification Task Performance}
Before evaluating self-explanation faithfulness, we validate whether the models used in the experiments could perform a classification task, for which they are required to generate self-explanations.

\begin{table}[t]
\centering
\resizebox{0.85\columnwidth}{!}{%
\begin{tabular}{lccc}
\hline
 & Sent140 & SNLI & AGNews \\ \hline
\footnotesize{Tulu-2 7B} &  &  &  \\
No Training & 0.737 & 0.760 & 0.750 \\
w/ Predictions & \textbf{0.896} & \textbf{0.911} & \textbf{0.904} \\
w/ Attribution & 0.804 & 0.685 & 0.532 \\
w/ Redaction & 0.780 & 0.706 & 0.743 \\
w/ Counterfactual & 0.700 & 0.740 & 0.634 \\ \hline
\footnotesize{Tulu-2 13B} &  &  &  \\
No Training & 0.712 & 0.814 & 0.815 \\
w/ Predictions & \textbf{0.901} & \textbf{0.918} & \textbf{0.905} \\
w/ Attribution & 0.788 & 0.653 & 0.597 \\
w/ Redaction & 0.773 & 0.703 & 0.807 \\
w/ Counterfactual & 0.795 & 0.698 & 0.818 \\ \hline
Chance Rate & 0.500 & 0.333 & 0.250 \\ \hline
\end{tabular}%
}
\caption{Classification accuracy before and after training. ``No-Training'' and ``w/ Predictions'' refer to the off-the-shelf models and those trained with ground-truth predictions, respectively. ``w/ Attribution'', ``w/ Redaction'' and ``w/ Counterfactual'' refer to models trained with self-explanations constructed for each style.}
\label{tab:classification-accuracy}
\end{table}

Table~\ref{tab:classification-accuracy} reports classification accuracy of the models before and after training, including those trained with the ground-truth predictions introduced in Section~\ref{subsec: Effects of Self-Explanation Training}.
The off-the-shelf Tulu-2 models score around 0.7 $\sim $ 0.8, while the prediction-trained models perform the best as expected, scoring around 0.9.
As for the models after training with the constructed self-explanations, we do not observe a significant drop in their prediction accuracies regardless of style, maintaining their classification performances sufficiently for faithfulness evaluation without serious catastrophic forgetting.

\section{Implementation Details}
\label{sec:appendix_implementation}
We implemented the codes for the experiments using Python v3.10.12, Py-Torch v2.5.1~\citep{paszke-2019-pytorch}, and Transformers v4.44.2~\cite{wolf-etal-2020-transformers}.
For word lemmatization, we used NLTK v3.9.1~\cite{bird-2009-nltk}.
Our study was conducted under the licenses and terms of the scientific artifacts.

We conducted the experiments with eight NVIDIA A100 (40GB) GPUs for dataset construction and training, and a single NVIDIA A100 (40GB) GPU for evaluation.
The construction of training datasets took approximately 21 GPU hours with Tulu-2 7B, and 30 GPU hours with Tulu-2 13B.
Training with instruction-tuning data combined with either ground-truth prediction responses or a self-explanation dataset takes approximately 8.19 GPU hours for Tulu-2 7B and 12.9 GPU hours for Tulu-2 13B.
Evaluation in each explanation style takes approximately 0,02  GPU hours for Tulu-2 7B, and 0.03 GPU hours for Tulu-2 13B, regardless of whether the model has been trained or not.

\begin{table*}[h!]
\resizebox{\textwidth}{!}{%
\begin{tabular}{llll}
\hline
 & \textbf{Input} & \textbf{Second Input} & \textbf{Ground Truth Predicton} \\ \hline
Sentiment140 & \begin{tabular}[c]{@{}l@{}}@cocodkr Not even superman\\ can save me now\end{tabular} & - & \begin{tabular}[c]{@{}l@{}}Positive\\ Negative \checkmark\end{tabular} \\ \hline
SNLI & \begin{tabular}[c]{@{}l@{}}A fisherman using a cellphone\\ on a boat.\end{tabular} & \begin{tabular}[c]{@{}l@{}}A fisherman is sleeping on his\\ boat.\end{tabular} & \begin{tabular}[c]{@{}l@{}}Entailment\\ Contradiction \checkmark\\ Neutral\end{tabular} \\ \hline
AGNews & \begin{tabular}[c]{@{}l@{}}Next space station crew to \\ launch\end{tabular} & - & \begin{tabular}[c]{@{}l@{}}World politics\\ Sports\\ Business\\ Science and technology \checkmark\end{tabular} \\ \hline
\end{tabular}%
}
\caption{Examples of each prediction dataset. ``Input'' refers to social networking posts in Sentiment140, premise sentences in SNLI, and news titles in AGNews, respectively. SNLI also includes hypothesis sentences as the second input.}
\label{tab:exampled_of_classification_datasets}
\end{table*}

\begin{table*}[h!]
\centering
\begin{tabular}{llrrr}
\hline
 & \textbf{Split }& \textbf{\# of Examples} & \begin{tabular}[c]{@{}r@{}}\textbf{Input}\\ \textbf{Avg. Length}\end{tabular} & \begin{tabular}[c]{@{}r@{}}\textbf{Second Input}\\ \textbf{Avg. Length}\end{tabular} \\ \hline
\multirow{2}{*}{Sentiment140} & Train & 50,000 & 13.17 & - \\
 & Test & 5,000 & 13.09 & - \\ \hline
\multirow{2}{*}{SNLI} & Train & 49,998 & 12.84 & 7.43 \\
 & Test & 4,998 & 13.88 & 7.53 \\ \hline
\multirow{2}{*}{AGNews} & Train & 50,000 & 6.78 & - \\
 & Test & 5,000 & 6.76 & - \\ \hline
\end{tabular}%
\caption{Statistics of the classification datasets used for our experiments.}
\label{tab:classification_dataset_statistics}
\end{table*}

\begin{table*}[htbp]
  \centering
  \begin{subtable}{\textwidth}
    \centering
    \resizebox{\textwidth}{!}{%
    \begin{tabular}{llccccccccccc}
    \hline
     &  & \multicolumn{3}{c}{Attribution} &  & \multicolumn{3}{c}{Redaction} &  & \multicolumn{3}{c}{Counterfactual} \\ \cline{3-5} \cline{7-9} \cline{11-13} 
     &  & Sent140 & SNLI & AGNews &  & Sent140 & SNLI & AGNews &  & Sent140 & SNLI & AGNews \\ \hline
    No Training &  & 4600 & 4925 & 4528 &  & 886 & 1330 & 1743 &  & 529 & 659 & 691 \\
    w/ Explanations &  & 4952 & 4992 & 4980 &  & 4706 & 4984 & 4978 &  & 4215 & 4834 & 4895 \\ \hline
    \end{tabular}%
    }
    \caption{Tulu-2 7B}
  \end{subtable}
  \hfill
  \vspace{0.2em}
  \\
  \begin{subtable}{\textwidth}
    \centering
    \resizebox{\textwidth}{!}{%
    \begin{tabular}{lllccccccccccc}
    \hline
    \multirow{2}{*}{} & \multirow{2}{*}{} &  & \multicolumn{3}{c}{Attribution} &  & \multicolumn{3}{c}{Redaction} &  & \multicolumn{3}{c}{Counterfactual} \\ \cline{4-6} \cline{8-10} \cline{12-14} 
     &  &  & Sent140 & SNLI & AGNews &  & Sent140 & SNLI & AGNews &  & Sent140 & SNLI & AGNews \\ \hline
    No Training & No Training &  & 4618 & 4865 & 4204 &  & 1291 & 685 & 2743 &  & 1072 & 536 & 1094 \\ \hline
    \multirow{3}{*}{w/ Attribution} & Sent140 &  & 4964 & 4773 & 4908 &  & 175 & - & - &  & 1069 & - & - \\
     & SNLI &  & 4861 & 4991 & 4901 &  & - & 1382 & - &  & - & 783 & - \\
     & AGNews &  & 4927 & 4971 & 4978 &  & - & - & 1 &  & - & - & 622 \\ \hline
    \multirow{3}{*}{w/ Redaction} & Sent140 &  & 3293 & - & - &  & 4719 & 3454 & 4599 &  & 96 & - & - \\
     & SNLI &  & - & 3371 & - &  & 4032 & 4983 & 4443 &  & - & 410 & - \\
     & AGNews &  & - & - & 4776 &  & 4197 & 4589 & 4972 &  & - & - & 30 \\ \hline
    \multirow{3}{*}{w/ Counterfactual} & Sent140 &  & 4793 & - & - &  & 617 & - & - &  & 4516 & 1579 & 3272 \\
     & SNLI &  & - & 4266 & - &  & - & 46 & - &  & 2096 & 4874 & 2254 \\
     & AGNews &  & - & - & 4683 &  & - & - & 1 &  & 3200 & 1253 & 4867 \\ \hline
    \end{tabular}%
    }
    \caption{Tulu-2 13B}
  \end{subtable}
  
  \caption{The number of evaluation instances used for faithfulness assessment of Tulu-2 7B and 13B. Instances in which generated self-explanations did not meet the requirements specified in the prompts were excluded from evaluation (see Section~\ref{subsec:evaluation}).}
  \label{tab:number_of_evaluated_instances}
\end{table*}

\begin{table*}[t!]
\centering
\begin{tabular}{lp{0.13\textwidth}p{0.65\textwidth}}
\hline
Dataset & Task & Prompt \\\hline\hline

Sentiment140 & Classification & Text: \{input\}\textbackslash n\textbackslash n Question: What would you classify the sentiment of the text as? The text can contain redacted words marked with [REDACTED]. Answer either `Positive' or `Negative' in a single word. Do not explain the answer.\\
 \rule{0pt}{3ex}  
 & Attribution (one-word) & Question: List the single most important word for determining the sentiment of the text, such that without this word the sentiment cannot be determined. Answer one word following `Answer:'. Do not explain the answer. \\
  \rule{0pt}{3ex}  
 & Attribution (multi-word)  & Question: List all and only the most important words for determining the sentiment of the text, such that without these words the sentiment cannot be determined. Answer in JSON format as \{``words'': [``word1'', ``word2'', ..., ``wordN'']\}.\\
 \rule{0pt}{3ex}  
 & Redaction (one-word)  & Question: Redact one word in the text by replacing the single most important word for determining the sentiment with '[REDACTED]'. Answer with the new redacted text following 'Answer:'. Do not explain your answer.\\
  \rule{0pt}{3ex}  
 & Redaction (multi-word)  & Question: Replace all and only the most important words in the text with '[REDACTED]', such as changing ``AAA BBB CCC'' into ``AAA [REDACTED] [REDACTED]''. Answer in JSON format as \{``redacted\_text'': ``<text with words replaced by [REDACTED]>''\}. \\
  \rule{0pt}{3ex}  
 & Counterfactual (one-word)  & Question: Edit the text by adding, removing, or replacing a single word so that the predicted sentiment changes from the current one. Do not use either '[REDACTED]' or the sentiment label itself. Answer with the new edited text following 'Answer:'. Do not explain your answer.\\
 \rule{0pt}{3ex}  
 & Counterfactual (multi-word)  & Question: Edit the text by adding, removing, or replacing words, making sure to change all and only the words necessary so that the predicted sentiment changes from the current one. Do not use either '[REDACTED]' or the sentiment label itself. Answer in JSON format as \{``edited\_text'': ``<text with exactly two words edited>''\}\\\hline
\end{tabular}
\caption{Prompt templates we use for Sentiment140 in the experiments. The placeholders of \{input\} is replaced with the appropriate strings for each instance.}
\label{tab:prompts_for_sentiment140}
\end{table*}

\begin{table*}[t!]
\centering
\begin{tabular}{lp{0.13\textwidth}p{0.65\textwidth}}
\hline
Dataset & Task & Prompt \\\hline\hline

SNLI & Classification & Sentence: \{input\}\textbackslash n\textbackslash n Question: Does this sentence imply that `\{second input\}'? The sentence can contain redacted words marked with [REDACTED]. Answer either `Yes', `No', or `Maybe' in a single word. Do not explain the answer. \\
 \rule{0pt}{3ex}  
 & Attribution (one-word) & Question: List the single most important word in the sentence, for determining the implication. Answer one word following 'Answer:'. Do not explain the answer.\\
  \rule{0pt}{3ex}  
 & Redaction (one-word) & Question: Redact one word in the sentence by replacing the single most important word for determining whether it entails `\{second input\}' with '[REDACTED]'. Answer with the new redacted sentence following 'Answer:'. Do not explain your answer.\\
  \rule{0pt}{3ex}  
 & Counterfactual (one-word) & Question: Edit the sentence by adding, removing, or replacing a single word so that the predicted NLI relationship to `\{second input\}' changes from the current one. Do not use either '[REDACTED]' or the NLI label itself. Answer with the new edited sentence following 'Answer:'. Do not explain your answer.\\\hline
 
 AGnews & Classification & Title: \{input\}\textbackslash n\textbackslash n Question: What label best describes this news title? The title can contain redacted words marked with [REDACTED]. Respond with one of the following single words: 'World', 'Sport', 'Business', or 'Tech'. Do not explain the answer.\\
 \rule{0pt}{3ex}  
 & Attribution (one-word) & Question: List the single most important word in the title for determining its topic label. Answer one word following 'Answer:'. Do not explain the answer.\\
  \rule{0pt}{3ex}  
 & Redaction (one-word) & Question: Redact one word in the title by replacing the single most important word for determining the topic label with '[REDACTED]'. Answer with the new redacted title following 'Answer:'. Do not explain your answer.\\
  \rule{0pt}{3ex}  
 & Counterfactual (one-word) & Question: Edit the title by adding, removing, or replacing a single word so that the predicted topic label changes from the current one. Do not use either '[REDACTED]' or the topic label itself. Answer with the new edited title following 'Answer:'. Do not explain your answer.\\\hline
\end{tabular}
\caption{Prompt templates we use for SNLI and AGNews in the experiments. The placeholders of \{input\} and \{second input\} are replaced with the appropriate strings for each instance.}
\label{tab:prompts_for_snli_agnews}
\end{table*}

\begin{table*}[t]
\resizebox{\textwidth}{!}{%
\centering
\begin{tabular}{p{0.13\textwidth}p{0.85\textwidth}}
\hline
Dataset & Prompt \\ \hline
Sentiment140 & You are given an English sentence with one redacted part, represented as {[}REDACTED{]}, and a target sentiment prediction ('Positive' or 'Negative'). Replace {[}REDACTED{]} with exactly one word that would make the completed sentence very likely to be predicted with the target sentiment. Keep the sentence natural and fluent, do not mention the sentiment label itself. Output only the replacement word. Do not explain the answer.\textbackslash{}n\textbackslash{}nSentence with redaction: \{redacted\_input\}\textbackslash{}nTarget label: \{target\_label\}\textbackslash{}nOutput word: \\ \hline
SNLI & You are given a premise–hypothesis pair in English. The premise contains one redacted part, represented as {[}REDACTED{]}, and a target NLI prediction ('Yes,' 'No,' or 'Maybe'). Replace {[}REDACTED{]} with exactly one word that would make the completed premise–hypothesis pair very likely to be predicted with the target answer. Keep both sentences natural and fluent, and do not mention the answer itself. Output only the replacement word. Do not explain the answer.\textbackslash{}n\textbackslash{}nPremise with redaction: \{redacted\_input\}\textbackslash{}nHypothesis: \{second\_input\}\textbackslash{}nTarget label: \{target\_label\}\textbackslash{}nOutput word: \\ \hline
AGNews & You are given an English news title with one redacted part, represented as {[}REDACTED{]}, and a target topic prediction ('World', 'Sport', 'Business', or 'Tech'). Replace {[}REDACTED{]} with exactly one word that would make the completed title very likely to be predicted with the target topic. Keep the title natural and fluent, and do not mention the topic label itself. Output only the replacement word. Do not explain the answer.\textbackslash{}n\textbackslash{}nTitle with redaction: \{redacted\_input\}\textbackslash{}nTarget label: \{target\_label\}\textbackslash{}nOutput word: \\ \hline
\end{tabular}%
}
\caption{Prompt templates we use for obtaining the word $w_{\bar{y}}$ during the construction of the counterfactual self-explanation datasets. The placeholders of \{redacted\_input\} and \{target\_label\} are replaced with the appropriate strings of the redacted input $x_{-w*}$ and $\bar{y}$, respectively, for each instance. In SNLI, \{second\_input\} is also replaced with adequate strings for each instance. See Section~\ref{sec: Self-Explanation Training} for the details.}
\label{tab:prompts_counterfactual_construction}
\end{table*}

\end{document}